\documentclass{article}

\usepackage[final,nonatbib]{nips_2018}

\usepackage[utf8]{inputenc} 
\usepackage[T1]{fontenc}    
\usepackage{hyperref}       
\usepackage{url}            
\usepackage{booktabs}       
\usepackage{amsfonts}       
\usepackage{nicefrac}       
\usepackage{microtype}      
\usepackage{color}
\usepackage{xcolor}
\usepackage{tcolorbox}
\usepackage{amsmath}
\usepackage{breqn}
\usepackage{float}
\usepackage{graphicx}
\usepackage{caption}
\usepackage{subcaption}
\usepackage{comment}
\usepackage{color}
\usepackage{etoolbox}
\usepackage{fancyhdr}
\newbool{inccomment}
\setbool{inccomment}{true}

\usepackage{amsmath,amssymb,amsfonts}
\usepackage{algorithmic}
\usepackage{graphicx}
\usepackage{textcomp}
\usepackage{xcolor}
\usepackage{soul}
\usepackage[ruled]{algorithm2e}
\newcommand{\yhr}[1]{\ifbool{inccomment}{{\color{blue}#1}}{}}
\newcommand{\ww}[1]{\ifbool{inccomment}{{\color{red}#1}}{}}
\definecolor{dave-fancy}{RGB}{200,100,0}
\newcommand{\dave}[1]{\ifbool{inccomment}{{\color{blue}#1}}{}}
\newcommand{\davenote}[1]{\ifbool{inccomment}{{\color{pink}#1}}{}}
\usepackage[normalem]{ulem}

\title{LEASGD: an Efficient and Privacy-Preserving Decentralized Algorithm for Distributed Learning}
\usepackage{times}
\usepackage{longtable}
\usepackage[utf8]{inputenc} 
\usepackage[T1]{fontenc}    
\usepackage{hyperref}       
\usepackage{url}            
\usepackage{booktabs}       
\usepackage{amsfonts}       
\usepackage{nicefrac}       
\usepackage{microtype}      
\usepackage{amsmath}
\usepackage{breqn}
\usepackage{float}
\usepackage{graphicx}
\usepackage{caption}
\usepackage{subcaption}
\usepackage{comment}
\usepackage{tabularx}  
\usepackage{etoolbox}

\setbool{inccomment}{true}
\usepackage{amsmath,amssymb,amsfonts}
\usepackage{algorithmic}
\usepackage{graphicx}
\usepackage{textcomp}
\usepackage{soul}
\usepackage{multirow}
\usepackage{enumitem}
\usepackage[ruled]{algorithm2e}

\newcommand{\leader}{leader\xspace}
\newcommand{\follower}{follower\xspace}
\newcommand{\leaderss}{leaders\xspace}
\newcommand{\followers}{followers\xspace}
\newcommand{\Leader}{Leader\xspace}
\newcommand{\Follower}{Follower\xspace}
\newcommand{\LEASGD}{LEASGD\xspace}
\newcommand{\davefixed}[1]{\ifbool{inccomment}{{\color{black}#1}}{}}
\newcommand{\yan}[1]{\ifbool{inccomment}{{\color{blue}#1}}{}}

\newtheorem{mypro}{Proposition}
\usepackage[normalem]{ulem}

\author{
  Hsin-Pai Cheng\thanks{Equal Contribution}\\
  ECE Department\\
  Duke University\\
  Durham, NC 27708 \\
  \texttt{hc218@duke.edu}
  \And
  Patrick Yu\footnotemark[1] \\
  Monta Vista High School\\
  Cupertino, CA 95014\\
  \texttt{pyu592@student.fuhsd.org} \\
  \AND
  Haojing Hu\footnotemark[1] \\
  Beihang University of \\
  Aeronautics and Astronautics\\
  Beijing, China \\
  \texttt{haojinghu@buaa.edu.cn} \\
  \And
  Feng Yan \\
  Computer Science\\ and Engineering\\
  University of Nevada, Reno\\
  Reno, NV 89557 \\
  \texttt{fyan@unr.edu} \\
  \And
  Shiyu Li \\
  Tsinhua University\\
  Beijing, China\\
  \texttt{shiyu.li@duke.edu} \\
  \And
  Hai Li \\
  ECE Department\\
  Duke University\\
  Durham, NC 27708 \\
  \texttt{hai.li@duke.edu} \\
  \And
  Yiran Chen \\
  ECE Department\\
  Duke University\\
  Durham, NC 27708 \\
  \texttt{yiran.chen@duke.edu} \\
}

\begin{document}

\maketitle

\renewcommand{\thefootnote}{\fnsymbol{footnote}}

\begin{abstract}
 Distributed learning systems have enabled training large-scale models over large amount of data in significantly shorter time. 
 In this paper, we focus on decentralized distributed deep learning systems and aim to achieve differential privacy with good convergence rate and low communication cost. 
 To achieve this goal, we propose a new learning algorithm LEASGD (Leader-Follower Elastic Averaging Stochastic Gradient Descent), which is driven by a novel Leader-Follower topology and a differential privacy model.
 We provide a theoretical analysis of the convergence rate and the trade-off between the performance and privacy in the private setting. 
  The experimental results show that LEASGD outperforms state-of-the-art decentralized learning algorithm DPSGD by achieving steadily lower loss within the same iterations and by reducing the communication cost by $30\%$. In addition, LEASGD spends less differential privacy budget and has higher final accuracy result than DPSGD under private setting. 
\end{abstract}

\section{Introduction}
\label{sec:intro}


With data explosion and ever-deeper neural network structures,
 distributed learning systems play an increasingly important role in training large-scale models with big training data sources \cite{Abadi2016TensorFlow}\cite{Dean2013Large}\cite{Xing2015Petuum}. 
Most distributed learning systems have centralized parameter server(s) to maintain a single global copy of the model and coordinate information among workers/clients. However, such system topology is vulnerable to privacy leakage because once the central server(s) is eavesdropped, information of the entire system can be exposed~\cite{agarwal2018cpsgd}.
Decentralized distributed learning systems are potentially more robust to the privacy as critical information such as training data, model weights, and the states of all workers can no longer be observed or controlled through a single point of the system~\cite{bellet2018}.

However, decentralized systems usually perform worse in convergence rate and are known to have higher communication cost.
In addition, most of the decentralized systems do not provide guarantees on differential privacy~\cite{NIPS2017_7117, yan2013distributed}.
Some recent works are proposed to solve the above problems.
For example, D-PSGD~\cite{NIPS2017_7117} focuses on improving communication efficiency and convergence rate of decentralized learning systems, but it is not differentially private. Bellet \textit{et al.} considers both decentralized design and differential privacy in their recent work~\cite{bellet2018}. However, it is based on a simple linear classification task, not a good representation of the modern neural networks, which have much more complex and deeper structures. Inspired by the above work, we aim at developing a generalized decentralized learning approach that has better applicability and can achieve differential privacy with good convergence rate and low communication cost. 

To this end, we propose LEASGD (\emph{Leader-Follower Elastic Averaging Stochastic Gradient Descent}) that provides differential privacy with improved convergence rate and communication efficiency.
To improve both the communication and training efficiency while also facilitate differential privacy, we first design a novel communication protocol that is driven by a dynamic \leader-\follower approach. 
The parameters are only transferred between the leader-follower pair, which significantly reduces the overall communication cost.
\LEASGD adopts the insight of the \emph{Elastic Averaging Stochastic Gradient Descent} (EASGD) \cite{zhang2015deep} by exerting the elastic force between the \leader and \follower at each update.
Inspired by~\cite{abadi2016deep}, we use \emph{momentum account} to quantize the privacy budget which provides a tighter bound of privacy budget $\epsilon$ than the classical \emph{Strong Composition Theorem} \cite{Dwork2010Boosting}. In addition, the convergence rate of \LEASGD is mathematically proved.



We evaluate LEASGD against the state-of-the-art approach D-PSGD~\cite{NIPS2017_7117} on three main aspects: the convergence rate, the communication cost, and the privacy level. The theoretical analysis shows \LEASGD converge faster than D-PSGD. 
Our real testbed experiments 
show \LEASGD achieves higher accuracy than D-PSGD in the non-private setting with the same communication cost and under the private setting with less privacy budget.




\section{Leader-Follower Elastic Averaging Stochastic Gradient Descent Algorithm and Privacy-preserving Scheme}
\label{sec:pro}
\subsection{Problem Setting}

We assume there are $m$ workers each with a set of local data $S_i$ and $i \in \{1,2......m\}$, which can only be accessed locally by worker $i$. 
Along the training process, each worker $i$ computes a parameter vector $w^i_t$ at each iteration $t$ to represent the learning outcomes and then computes the corresponding loss function $f^i_t(w^i)=l(w^i_t,x^i_t,y^i_t)$ with the input $x^i_t$ and given labels $y^i_t$. 
After learning from the data, each worker has two ways to contribute to the global learning progress: 1) Update its model parameters by local gradient descent; 2) Communicate with other workers to update each other's model parameters. 
We define a communication interval $\tau$ to represent how many iterations between each update in our learning algorithm. 
When the training process is done, each worker has its own variation of the same model (i.e., performing the same task but with different trained model parameters $w^i$). 
Given each worker has its own local version of the model, it is necessity to assemble all local models unbiasedly by averaging the loss function. We formulate it as an optimization problem as follows:  

\vspace{-4mm}
\begin{equation}
    \label{eq:objective_func}
        w^*=\{w^1,...,w^m\}
\end{equation}
\vspace{-3mm}
\begin{equation}
    \label{objective_func}
    \begin{split}
        \mathop{argmin}\limits_{w^i\in\Omega}\overline{F}(w,T)=\frac{1}{m}\sum_{i=1}^{m}f^i_T(w^i), 
        s.t. \Omega\subseteq\mathbb{R}^n\,and\,T\in\mathbb{R}
    \end{split}
\end{equation}
where $T$ presents the predefined number of iterations in $\tau$.



\subsection{ Decentralized \Leader-\Follower Topology}
\begin{figure*}[t]
    \vspace{-3mm}
	\centering
	\includegraphics[width=1\textwidth]{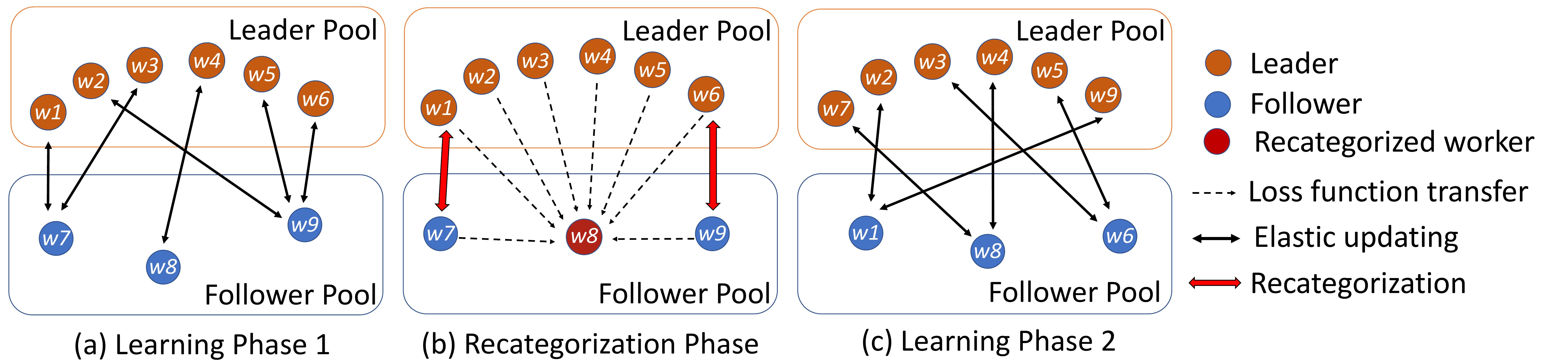}
    \vspace{-4mm}
	\caption{The dynamic \Leader-\Follower topology. (a) shows the structure of \leader pool and \follower pool. (b) shows a recategorization phase where one of the workers is elected as recategorized worker and gathers the latest loss function values from all other workers. (c) shows the new structure of \leader pool and \follower pool after recategorization (note the randomization used for avoiding over-fitting). 
	}
	\label{fig:overview}
	\vspace{-4mm}
\end{figure*}

To support the decentralized design, we categorize all workers into two worker pools: \leader pool with workers of lower loss function values and \follower pool with workers of higher loss function values, see Figure~\ref{fig:overview}(a). The leader pools are always larger than the follower to guarantee the sufficient pull power.
The core idea is to let \leaderss to pull \follower so that better performing workers (\leaderss) can guide the 
\followers in the right direction to improve the learning.
Specifically, we use an elastic updating rule to regulate the learning updates in each leader-follower pair as follows:

\vspace{-6mm}

\begin{equation}
\label{elastic_update}
    \begin{split}
        w^i_{t+1}=w^i_{t}-\eta g^i_t+\eta\rho(w^f_{t}-w^i_{t}) \quad and \quad
        w^f_{t+1}=w^f_{t}-\eta g^f_t+\eta\rho(w^i_{t}-w^f_{t})
    \end{split}
\end{equation}


We use $i$ to denote a leader; $f$ is a follower; $k$ is the categorization interval; $\rho$ is elastic factor; $g$ is gradient; and $\eta$ is learning rate. Given learning is a dynamic process, the two worker pools are dynamically updated based on the learning progress. The pools are recategorized each $k\tau$ time interval. 
This protocol enables the convergence rate of our algorithm to have a limited upper bound. 
To avoid over-fitting to one worker's model during the training process, we add the L2-normalization on the training loss function.
We also randomly pair the leaders and followers after each learning update to 
avoid one follower's model having excessive influence on others. 
This randomization mechanism also benefits the privacy-preserving as randomized communication can confuse the attacker and make it more difficult to trace the information source. 
An asynchronous version can be derived by setting the number of wake up iterations for different workers according to a Poisson Stochastic Process with different arrival rate based on local clock time of each worker, 
see Supplementary for more details.



\subsection{Privacy-preserving Scheme}
The general idea to preserve differential privacy is to add noise on the output of the algorithm and the noise scale is based on the sensitivity of the output function as defined in \cite{dwork2006}.
Note that for different input data, equation \ref{elastic_update} only differs in the gradient $g_t^i$ part. In other words, the sensitivity of the updating rule of LEASGD is the same as the gradient $g_t^i$. Thus we use the similar scheme as the DP-SGD algorithm \cite{abadi2016deep}.
To limit the sensitivity of gradient, we clip the gradient into a constant $C$ as 
$\overline{g}^i_t=g_t^i/max(1,\frac{\parallel g^i_t\parallel_2}{C})$. Then, we add Gaussian noise on the clipped gradient
\begin{equation}
\label{gaussian_n}
   \begin{split}
        \tilde{g}^i_t=\overline{g}^i_t+\mathcal{N}(0, \sigma_2^2C^2)
    \end{split}
\end{equation}
By using $\tilde{g}^i_t$
to replace $g^i_t$ in equation \ref{elastic_update}, we obtain the differential-privacy preserving scheme of LEASGD as:
\begin{equation}
\label{private_elastic_rule}
    \begin{split}
        \tilde{w}^i_{t+1}=\tilde{w}^i_{t}-\eta \tilde{g}^i_t+\eta\rho(\tilde{w}^f_{t}-\tilde{w}^i_{t}) \quad and \quad
        \tilde{w}^f_{t+1}=\tilde{w}^f_{t}-\eta \tilde{g}^f_t+\eta\rho(\tilde{w}^i_{t}-\tilde{w}^f_{t})
    \end{split}
\end{equation}
When we choose the variance of Gaussian noise $\sigma_2=\frac{\sqrt{2ln(1.25/\delta)}}{\epsilon}$, we ensure that each communication step of LEAGSD is ($\epsilon,\delta$)-DP. Using the property of DP-mechanism in \cite{Dwork2010Boosting}, the composition of a series of DP-mechanisms remains DP, which guarantees that for each worker $i$, its training algorithm $\mathcal{M}_i$ at each iteration is DP. 
\section{Analysis}
\label{sec:ana}
{\bf Convergence Rate Analysis.}
In this section, we provide a convergence rate analysis for synchronous LEASGD in a strongly-convex case and also compare it with the D-PSGD~\cite{NIPS2017_7117} theoretically. 




We define that
\begin{equation}
    \label{d_t}
    \begin{split}
        d_t=\frac{E\sum_{i=1}^{p}\parallel w^i_t-w^*\parallel^2+E\parallel w^f_t-w^*\parallel^2}{p+1}
    \end{split}
\end{equation}

\vspace{-3mm}

and the result of convergence rate of $d_t$ is as follow.

\begin{mypro}{(Convergence rate of d_t)}
If $0\le\eta\le\frac{2(1-\beta)}{\mu+L}$,$0\le\alpha=\eta\rho<1$,$0\le\beta=p\alpha<1$, then we obtain the convergence of $d_t$\\
\begin{equation}\label{eq:rate}
    \begin{split}
        d_{t}\le h^t d_0+(c_0-\frac{\eta^2\sigma_1^2}{\gamma})(1-\gamma)^t(1-(\frac{p}{p+1})^t)
        +\eta^2\sigma_1^2\frac{1-h^t}{\gamma},\\
        where\  0<h=\frac{p(1-\gamma)}{p+1}<1,k=\frac{1-\gamma}{p+1},\gamma=2\eta\frac{\mu L}{\mu+L}, \ c_0=\mathop{max}\limits_{i=1,...,p,f}\parallel w^i_0-w^*\parallel^2
    \end{split}
\end{equation}
\end{mypro}

\vspace{-2mm}
This proposition implies that the average gap between all workers and optimum in a subsystem includes three parts, which could also be applied to the whole system. 
If we simply ignore the influence of the inherent noise on the gradient and extend the $t\to\infty$, we can easily obtain an purely exponential decline of the gap, that is $E[d_{t+1}]\le hE[d_t]$. Note that the shrink factor $h$ is negatively correlated to the $p$, which implies that, when our system is operating in a strongly convex setting, the larger worker scale can correspondlingly result in a faster convergence rate of the system. 
The convergence rate of D-PSGD \cite{NIPS2017_7117} is $O(1/[(p+1)t])$ with our denotation in the strongly-convex setting. Compared with our $O(h^t)$ rate, the convergence rate of D-PSGD is relatively slower when we extend the $t\to\infty$. Detailed proof can be found in Supplementary material.

{\bf Privacy trade-off analysis.} 
Following the convergence rate analysis above, we obtain the modified convergence rate of $d_t$ by adding the extra noise. We assume the DP noise is independent of the inherent noise. 
Thus, the variance of the composed noise is the sum of the two independent noise variances and it satisfies $\sigma^2<\sigma_1^2+C^2\sigma_2^2$. Finally, by replacing the $\sigma_1^2$ with $\sigma_1^2+C^2\sigma_2^2$ in Proposition 1, we obtain the convergence rate in the private setting. 
The extra trade-off can be formulated as $\frac{\eta^2C^2\sigma_2^2}{\gamma}$ when $t\to\infty$. Note that this trade-off remains the same when $p$ grows. It implies our algorithm has a stable scalability when applied in the private setting.
\section{Experimental Evaluation}
\label{sec:exp}

We perform experimental evaluation using a 3-layer Multi-layer Perceptron (MLP) and 3-layer CNN with MNIST and CIFAR-10, respectively, running on a cluster of 15 servers each equipped with 4 NVIDIA Tesla P100 GPUs and the servers are connected with 100Gb/s Intel Omni-Path fabric.



\begin{figure*}[t]
	\vspace{-12pt}
	\centering
	\includegraphics[width=0.45\textwidth]{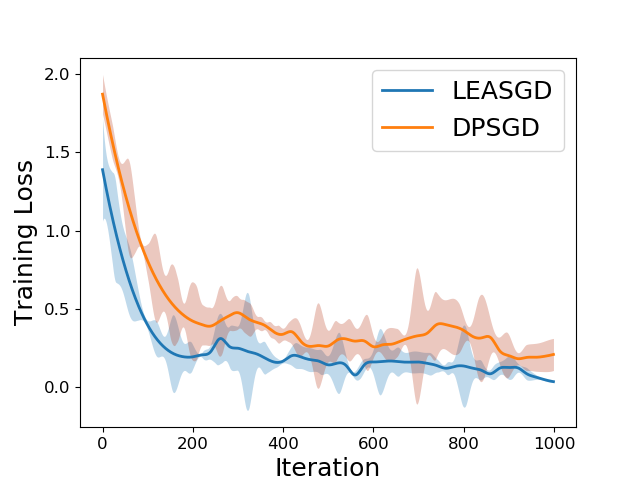}
	\includegraphics[width=0.45\textwidth]{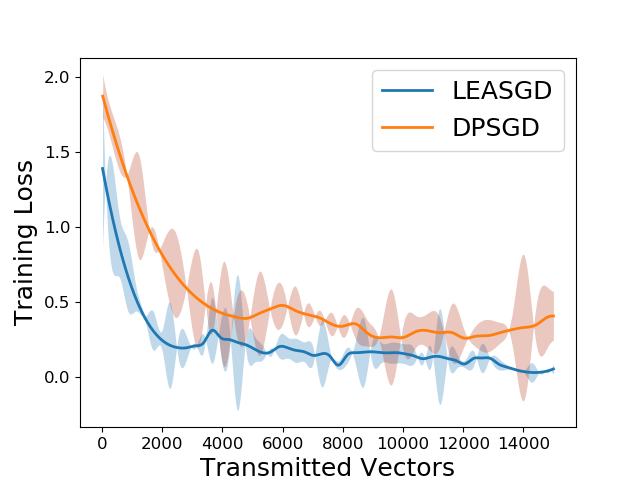}
	
\begin{minipage}[t]{.45\linewidth}
\centering
\vspace{-8pt}
\subcaption{}
\end{minipage}%
\begin{minipage}[t]{.45\linewidth}
\centering
\vspace{-8pt}
\subcaption{}
\end{minipage}%



\vspace{-3mm}
	\caption{(a). Training Loss vs. iteration on MNIST (b). Training Loss vs. number of transmitted vectors on MNIST}
	\label{fig:acc_overall}
	\vspace{-3mm}
\end{figure*}

{\bf Non-private setting comparison.} To quantitatively evaluate the communication cost, we track the average training loss of all workers by comparing between the proposed LEASGD and D-PSGD in terms of the number of iterations and transmitted vectors, see Figure \ref{fig:acc_overall}. Theoretically, in each iteration, \LEASGD has less transmitted vectors than D-PSGD.
Figure \ref{fig:acc_overall}(a) shows that \LEASGD converges faster than D-PSGD at the beginning of the training process and also achieves a lower loss function at the end. Figure \ref{fig:acc_overall}(b) shows \LEASGD outperforms D-PSGD in terms of the communication efficiency, i.e., with the same transmitted vectors, LEASGD achieves better training loss.

{\bf Differential private comparison.} In the private setting, we use the \emph{momentum account} to compute the totally spent $\epsilon$ and the adding noise scales are the same for two algorithms. As shown in Table 1, 
\LEASGD achieves better accuracy with less $\epsilon$ than D-PSGD. More importantly, the final accuracy of our algorithm does not vary greatly when the worker scale $m$ increases. We believe this result benefits from two attributes of \LEASGD: 1) the DP noise helps improve the accuracy by encouraging space exploration and helping workers trapped in local optimum to get out \cite{neelakantan2015adding}; 2) the great scalability that prevents DP noise from accumulating when the worker scale expands.

\begin{table}[bh!]
\centering
\begin{tabular}{ccccccccc} 
  \toprule
 &\multicolumn{4}{c}{m=5}&\multicolumn{4}{c}{m=15}\\[0.5ex] 
 \cmidrule(r){2-5}\cmidrule(r){6-9}
   & \multicolumn{2}{c}{ours}   & \multicolumn{2}{c}{D-PSGD} & \multicolumn{2}{c}{ours} & \multicolumn{2}{c}{D-PSGD}\\ 

 \hline
 
  & Accu. & Total $\epsilon$  & Accu. & Total $\epsilon$ & Accu. & Total $\epsilon$ & Accu. & Total $\epsilon$\\ 
 \hline
 
MNIST & \textbf{0.97} & \textbf{4.183} & 0.97 & 4.505
& \textbf{0.97} & \textbf{4.651} & 0.95 & 4.843\\
 CIFAR-10 & \textbf{0.74} & \textbf{4.651} & 0.71 & 4.925
 & \textbf{0.72} & \textbf{4.116} & 0.68 & 4.56\\

    \bottomrule
\end{tabular}
\caption{Private setting result of $\epsilon$ and accuracy\label{long}}
\label{table:1}
\end{table}

\section{Acknowledgement}
This work is supported in part by the following grants: National Science Foundation CCF-1756013, IIS-1838024, 1717657 and Air Force Research Laboratory FA8750-18-2-0057. 
\medskip

\small

\bibliographystyle{unsrt}
\bibliography{nips_2018}

\end{document}